\documentclass[runningheads]{llncs}

\usepackage{graphicx}
\usepackage{amsmath,amssymb}
\usepackage{booktabs}
\usepackage{multirow}
\usepackage{algorithm}
\usepackage{algorithmic}
\usepackage{xcolor}
\usepackage{cite}
\usepackage{hyperref}

\hypersetup{
    colorlinks=true,
    linkcolor=blue,
    citecolor=blue,
    urlcolor=blue,
}

\begin{document}
\title{FinSafeGuard: A Quality-Driven Synthetic Dataset for Financial AI Safety}
\titlerunning{FinSafeGuard}

\author{%
Joseph Thomas Thacil\inst{1} \and
Reetu Raj Harsh\inst{1} \and
Krati Saxena\inst{1} \and
Bhaskarjit Sarmah\inst{1}\thanks{Corresponding author: \email{bhaskarjit.sarmah@domyn.com}} \and
Stefano Pasquali\inst{2}%
}
\authorrunning{Thacil et al.}
\institute{%
Domyn, India\\
\email{josephthomas.thachil@domyn.com, reeturaj.harsh@domyn.com, krati.saxena@domyn.com, bhaskarjit.sarmah@domyn.com}
\and
Domyn, USA\\
\email{stefano.pasquali@domyn.com}%
}

\maketitle

\begin{abstract}
Generating high-quality synthetic datasets at scale remains a fundamental challenge for training specialized AI systems in data-scarce domains. We present a comprehensive methodology that produces 14.3 million quality-filtered synthetic training examples from a structured taxonomy of 632,000 scenario specifications. Our four-stage pipeline addresses critical challenges in synthetic data quality, diversity, and realism through: (1) systematic pre-generation research establishing optimal generation parameters via diminishing returns analysis, (2) conditional auto-evolution mechanisms that remediate low-diversity batches during generation, (3) a two-layer semantic deduplication architecture combining lexical and embedding-based methods, and (4) LLM-as-Judge quality evaluation filtering samples based on behavioral correctness and tuple fidelity. Empirical analysis across three domains (Banking, Insurance, Financial Services) demonstrates domain-specific deduplication patterns ranging from 9.9\% to 40.6\%, with output generation modes exhibiting highest redundancy. Our batch size optimization reveals mode-specific diversity saturation thresholds, enabling principled allocation strategies. The complete pipeline reduces 26.4M raw samples to 14.3M high-confidence examples through merge (20.9\% reduction), semantic deduplication (18.7\% reduction), and quality-based filtering (15.9\% reduction), providing a reproducible framework for domain-specific synthetic data generation with rigorous quality control.

\textbf{Keywords:} Synthetic data generation $\cdot$ Quality evaluation $\cdot$ Diversity metrics $\cdot$ Semantic deduplication $\cdot$ LLM-as-Judge $\cdot$ Financial AI safety
\end{abstract}

\section{Introduction}

Large language models have demonstrated remarkable capabilities across diverse tasks, yet their deployment in specialized domains requires training data that captures domain-specific risks, edge cases, and regulatory constraints. Financial institutions face unique challenges: real-world safety incidents are sparse, sensitive customer data cannot be shared, and regulatory requirements demand comprehensive coverage of potential failure modes \cite{bai2022constitutional,ouyang2022training}.

Synthetic data generation offers a promising solution, enabling creation of large-scale training datasets without privacy concerns or data scarcity limitations \cite{xu2023wizardlm,gunasekar2023textbooks}. However, naive scaling---simply generating millions of synthetic examples---produces datasets plagued by redundancy, low diversity, and quality inconsistencies that degrade downstream model performance \cite{zhou2023lima,lee2022deduplicating}.

This work addresses three fundamental questions in large-scale synthetic data generation:

\begin{enumerate}
\item \textbf{Batch Size Optimization:} How many samples should be generated per scenario specification before diversity saturates and duplicates emerge?
\item \textbf{Quality-Diversity Tradeoff:} How can we systematically measure and enforce both quality and diversity throughout the generation pipeline?
\item \textbf{Scalable Deduplication:} How can we efficiently detect and remove semantic duplicates across millions of generated samples?
\end{enumerate}

\subsection{Our Approach}

We present a four-stage pipeline that produces 14.3 million quality-filtered synthetic examples for financial AI safety applications across Banking, Financial Services, and Insurance domains. Our methodology combines:

\paragraph{Taxonomy-Driven Generation.} A comprehensive safety taxonomy expanded into 632,000 unique scenario tuples across 9 dimensions (domain, category, user sophistication, risk level, framing, channel, attack vectors, generation mode, safety label).

\paragraph{Pre-Generation Research.} Systematic experiments establishing optimal data generation process (DGP) parameters, mode-specific batch sizes, and validated diversity metrics before full-scale generation.

\paragraph{Conditional Auto-Evolution.} A chained mechanism that automatically detects low-diversity batches during generation and applies Evol-Instruct \cite{xu2023wizardlm} operators to remediate them, ensuring consistent quality without post-hoc filtering.

\paragraph{Two-Layer Deduplication.} Combining lexical (LSH + SimHash) and semantic (FAISS + MiniLM) methods to remove duplicates at sub-quadratic complexity, with domain-specific deduplication rates ranging from 9.9\% to 40.6\%.

\paragraph{LLM-as-Judge Quality Evaluation.} A two-stage assessment framework evaluating behavioral correctness and tuple fidelity, filtering samples with low-confidence labels to produce the final 14.3M training corpus.

\subsection{Contributions}

\begin{enumerate}
\item \textbf{Systematic Pre-Generation Research:} Empirical analysis of batch size vs. diversity tradeoffs, revealing mode-specific saturation thresholds (input: 20, output: 8, multi-turn: 3) through diminishing returns curves.

\item \textbf{Validated Diversity Scoring:} A 5-component diversity metric (uniqueness, length variety, pattern diversity, vocabulary richness, semantic spread) with Spearman correlations $\rho = 0.58$--$0.82$ against human judgments across 4,500 batches.

\item \textbf{Conditional Auto-Evolution:} Chained auto-evolve mechanism applying Evol-Instruct operators only to low-diversity batches (score $< 70$), reducing raw output by 20.9\% while preserving high-quality samples.

\item \textbf{Domain-Specific Deduplication Analysis:} Detailed characterization of redundancy patterns showing Banking output mode exhibits 40.6\% duplication (highest) while Financial Services input shows 9.9\% (lowest).

\item \textbf{Reproducible Pipeline:} Complete four-stage methodology from 26.4M raw samples to 14.3M quality-filtered examples, with total reduction of 45.8\% through principled quality control.
\end{enumerate}

\subsection{Paper Organization}

Section~\ref{sec:related} surveys related work. Section~\ref{sec:taxonomy} describes the taxonomy structure. Section~\ref{sec:pregeneration} presents pre-generation research. Section~\ref{sec:generation} details the data generation pipeline. Section~\ref{sec:deduplication} explains the two-layer deduplication architecture. Section~\ref{sec:quality} describes LLM-as-Judge evaluation. Section~\ref{sec:limitations} discusses limitations. Section~\ref{sec:conclusion} concludes.

\section{Related Work}
\label{sec:related}

\paragraph{Synthetic Data Generation.} Recent work has demonstrated the effectiveness of synthetic data for training language models. WizardLM \cite{xu2023wizardlm} introduced Evol-Instruct for instruction complexity enhancement. Textbooks Are All You Need \cite{gunasekar2023textbooks} showed that high-quality synthetic textbooks can match real data. Phi-3 \cite{abdin2024phi3} and Orca 2 \cite{mitra2023orca2} scaled synthetic data generation with careful quality control. Our work extends these approaches with systematic pre-generation research and domain-specific deduplication.

\paragraph{Data Quality and Diversity.} LIMA \cite{zhou2023lima} demonstrated that 1,000 carefully curated examples outperform datasets orders of magnitude larger, emphasizing quality over quantity. Our diversity scoring system builds on these insights, providing quantitative metrics for batch quality assessment during generation.

\paragraph{Deduplication Methods.} Lee et al. \cite{lee2022deduplicating} showed that deduplication significantly improves language model quality. SemDeDup \cite{abbas2023semdedup} introduced embedding-based semantic deduplication. Our two-layer architecture combines lexical and semantic methods, achieving sub-quadratic complexity through LSH and FAISS indexing.

\paragraph{AI Safety and Alignment.} Constitutional AI \cite{bai2022constitutional} and RLHF \cite{ouyang2022training} established frameworks for aligning language models with human values. Red-teaming approaches \cite{perez2022red,ganguli2022red} generate adversarial examples to identify model failures. Our work applies these principles to domain-specific safety in financial services.

\section{Financial Safety Taxonomy}
\label{sec:taxonomy}

Our synthetic data generation is driven by a comprehensive financial safety taxonomy covering Banking, Financial Services, and Insurance (BFSI) domains. The taxonomy defines 15 risk categories spanning financial integrity, regulatory compliance, fraud prevention, and domain-specific vulnerabilities.

\subsection{Taxonomy Structure}

Each scenario is specified by a 9-dimensional tuple:
\begin{equation}
Z = (d, t, u, k, f, ch, a, m, y)
\end{equation}

\noindent where:
\begin{itemize}
\item $d \in \{\text{Banking, Insurance, Financial Services}\}$: Domain
\item $t$: Taxonomy node (15 categories, 47 sub-categories)
\item $u \in \{\text{Novice, Intermediate, Expert}\}$: User sophistication
\item $k \in \{1, 2, 3\}$: Risk severity level
\item $f \in \{\text{Direct, Indirect, Hypothetical}\}$: Framing style
\item $ch \in \{\text{Chat, Email, Voice, Form}\}$: Interaction channel
\item $a$: Binary flags for 8 attack vectors (social engineering, jailbreak, etc.)
\item $m \in \{\text{Input, Output, Multi-Turn}\}$: Generation mode
\item $y \in \{\text{Safe, Unsafe}\}$: Target safety label
\end{itemize}

This factorization yields approximately 632,000 unique tuples after filtering implausible combinations through a structural causal model encoding realistic dependencies.

\subsection{Risk Categories}

The 15 risk categories include: (1) Financial Integrity \& Accuracy, (2) Regulatory Compliance, (3) Fraud \& AML Evasion, (4) Unauthorized Trading Advice, (5) Insurance Fraud Coaching, (6) Data Privacy Violations, (7) Market Manipulation, (8) Misrepresentation of Products, (9) Discriminatory Practices, (10) Cybersecurity Vulnerabilities, (11) Conflicts of Interest, (12) Inadequate Risk Disclosure, (13) Churning \& Excessive Trading, (14) Ponzi/Pyramid Schemes, and (15) Regulatory Arbitrage.

\section{Pre-Generation Research}
\label{sec:pregeneration}

Before scaling to full dataset generation across 632K tuples, we conduct systematic research to establish optimal parameters for the data generation process (DGP), batch size per tuple, and diversity scoring thresholds.

\subsection{Data Generation Process Design}

We formalize synthetic data generation as a factorized generative model. Let $Z_{\text{scenario}} = (d, t, u, k, f, ch, a)$ denote the 7 scenario dimensions of the full tuple (Section~\ref{sec:taxonomy}), with generation mode $m$ and target label $y$ treated as conditioning parameters rather than latent variables. Each sample is produced by:

\begin{align}
Z_{\text{scenario}} &\sim P(Z_{\text{scenario}}) \\
x &\sim P(x \mid Z_{\text{scenario}}, m) \\
(y, c, s) &\sim P(y, c, s \mid x, Z_{\text{scenario}})
\end{align}

\noindent where $x$ is the generated prompt/conversation, $y$ is the safety label, $c$ is the taxonomy category, and $s$ is the severity level. The LLM approximates $P(x \mid Z_{\text{scenario}}, m)$, while labels are either specified by the tuple (for targeted generation) or judged post-hoc.

\begin{figure}[t]
\centering
\includegraphics[width=0.85\textwidth]{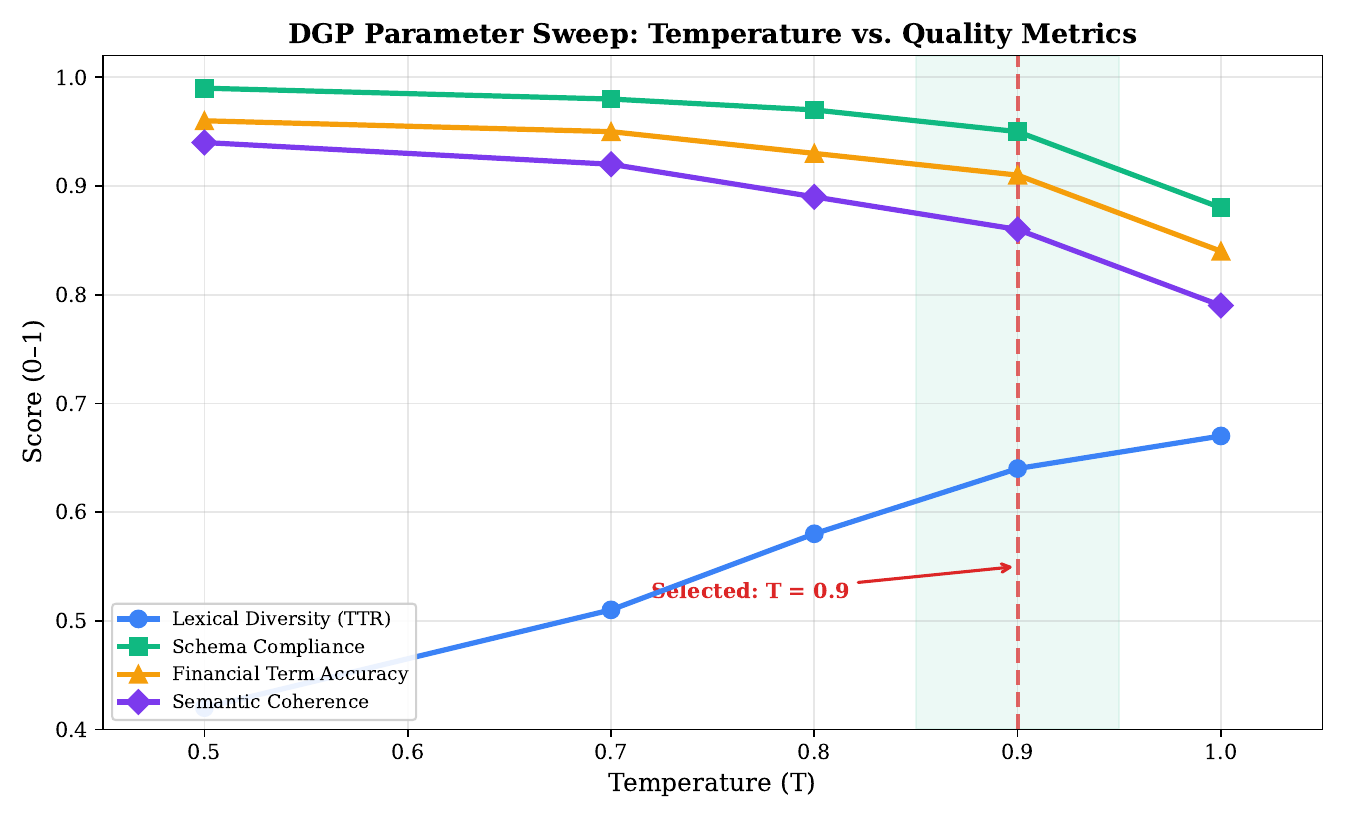}
\caption{Impact of temperature on generation quality metrics. Higher temperatures increase lexical diversity but reduce schema compliance beyond $T > 0.9$. The production pipeline uses $T = 0.9$ as the optimal balance.}
\label{fig:dgp_temperature}
\end{figure}

\paragraph{Temperature Sweep.} We evaluate generation quality across temperature settings $T \in \{0.5, 0.7, 0.8, 0.9, 1.0\}$, measuring lexical diversity (Type-Token Ratio), semantic coherence, financial terminology accuracy, and schema compliance rate. Figure~\ref{fig:dgp_temperature} shows that $T = 0.9$ provides optimal balance between creativity and coherence.

\subsection{Batch Size Optimization}

A \textit{batch} is the set of outputs generated from the same fixed input tuple in a single run. The batch size $N$ determines how many outputs we generate per tuple before moving to the next. This is critical because for any fixed input $X$, the LLM has finite conditional entropy $H(Y|X)$---it can only produce so many meaningfully different responses.

\paragraph{Theoretical Foundation.} Define a diversity functional $D(\cdot)$ measuring variety in a set of outputs. The optimal batch size $N^*$ is the smallest $N$ such that:
\begin{equation}
\mathbb{E}[D(N+1) - D(N)] \leq \epsilon
\end{equation}

\noindent where $\epsilon$ is our tolerance threshold. Diversity growth is sublinear because the LLM's conditional output space is bounded. We model this as:
\begin{equation}
D(n) = D_{\max} \cdot \left(1 - e^{-\lambda n}\right)
\end{equation}

\noindent where $D_{\max}$ is the asymptotic diversity and $\lambda$ is the mode-specific diversity growth rate. This idealized model captures the saturation behavior: diversity grows rapidly at small $N$ then plateaus as the LLM's conditional output space is exhausted. Higher-complexity modes saturate faster because generation constraints---coherence across turns, response-query consistency---naturally limit the space of meaningfully different outputs.

\paragraph{Experimental Setup.} We sample 100 random tuples and generate outputs at $N \in \{5, 10, 15, 20, 50, 100\}$ samples per tuple. For each configuration, we measure per-batch diversity, within-tuple similarity, cross-tuple overlap, and generation efficiency.

\begin{figure}[t]
\centering
\includegraphics[width=\textwidth]{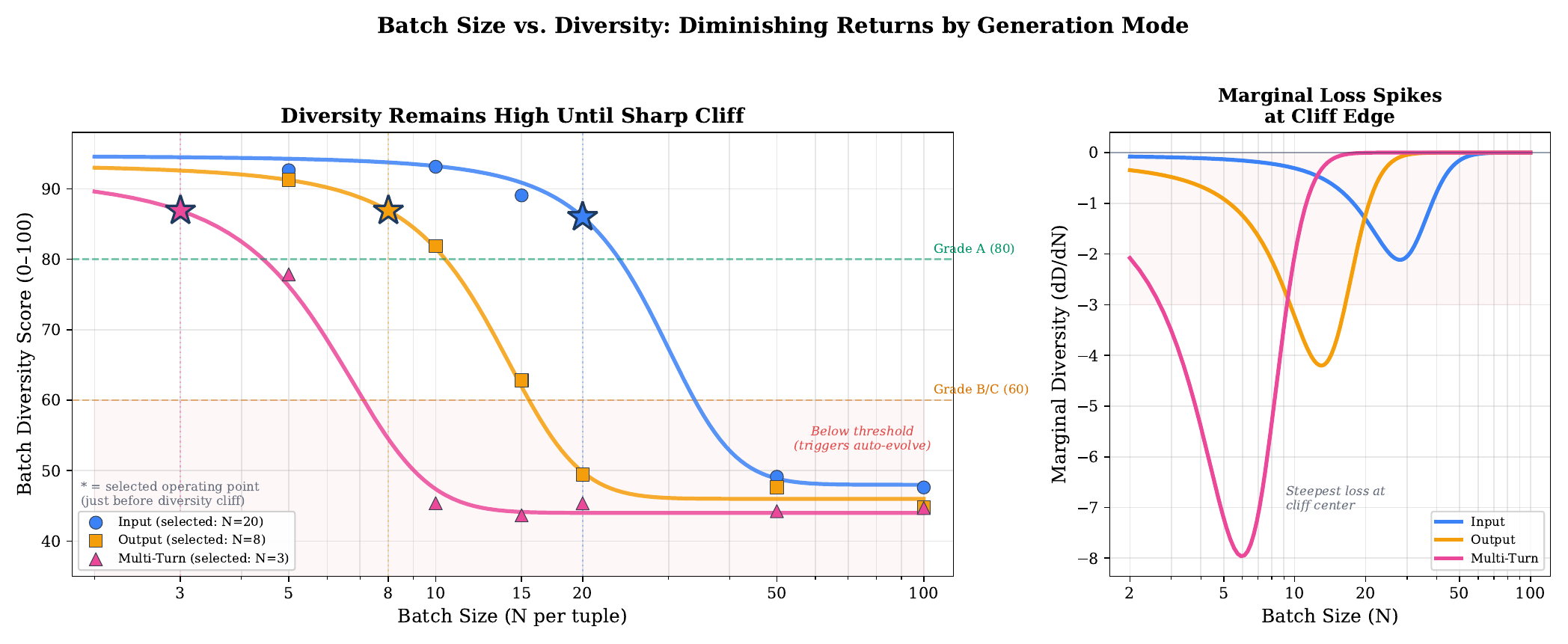}
\caption{Batch diversity score as a function of batch size per tuple (log scale). Diversity remains high ($>90$) up to each mode's capacity threshold, then drops as the LLM exhausts its conditional output space. Multi-turn saturates earliest ($N \approx 5$), output at $N \approx 12$, and input at $N \approx 25$. Selected batch sizes (stars) sit just before the empirical drop-off. Right panel shows marginal diversity loss, which spikes sharply at the transition point.}
\label{fig:batch_diversity}
\end{figure}

\paragraph{Diminishing Returns Analysis.} In practice, the empirical diversity curves (Figure~\ref{fig:batch_diversity}) exhibit a sharper transition than the smooth saturation of Equation~(5): diversity remains high ($>90$) up to mode-specific thresholds, then drops off as near-duplicate rates spike---a cliff-like behavior that the idealized exponential model understates. This motivates selecting batch sizes conservatively, just before the empirical drop-off. The marginal diversity loss follows:
\begin{equation}
\text{MarginalDiversity}(N) = \frac{\partial D}{\partial N} \propto -N^{-\alpha}, \quad \alpha \approx 0.6
\end{equation}

\paragraph{Mode-Specific Batch Sizes.} Based on these experiments, we select:
\begin{itemize}
\item \textbf{Input mode:} 20 samples/tuple (15 unsafe + 5 safe)
\item \textbf{Output mode:} 8 samples/tuple (5 unsafe + 3 safe)
\item \textbf{Multi-turn mode:} 3 samples/tuple (2 unsafe + 1 safe)
\end{itemize}

\subsection{Diversity Scoring Validation}

We develop a 5-component diversity scoring system (100 points total) that evaluates batch quality during generation. Table~\ref{tab:diversity_metrics} summarizes the components.

\begin{table}[t]
\centering
\caption{Five-Component Diversity Scoring System (100 points)}
\label{tab:diversity_metrics}
\begin{tabular}{@{}lccl@{}}
\toprule
\textbf{Component} & \textbf{Points} & \textbf{Method} & \textbf{Detects} \\
\midrule
Uniqueness & 0--30 & MD5 + SimHash & Exact and near-duplicates \\
Length Variety & 0--20 & CV $= \sigma / \mu$ & Template-based generation \\
Pattern Diversity & 0--20 & Multi-position intro & Repetitive openings \\
Vocabulary Richness & 0--15 & Type-Token Ratio & Vocabulary collapse \\
Semantic Spread & 0--15 & Embedding distance & Semantic similarity \\
\bottomrule
\end{tabular}
\end{table}

\paragraph{Validation Against Human Ratings.} We validate these metrics against human diversity ratings on 4,500 batches. Table~\ref{tab:diversity_validation} reports Spearman rank correlations.

\begin{table}[t]
\centering
\caption{Diversity Metric Validation Against Human Ratings}
\label{tab:diversity_validation}
\begin{tabular}{@{}lcc@{}}
\toprule
\textbf{Metric} & \textbf{Spearman $\rho$} & \textbf{Weight} \\
\midrule
Uniqueness (MD5+SimHash) & 0.82 & 30\% \\
Length Variety (CV) & 0.61 & 20\% \\
Pattern Diversity & 0.73 & 20\% \\
Vocabulary Richness (TTR) & 0.58 & 15\% \\
Semantic Spread (Embeddings) & 0.77 & 15\% \\
\bottomrule
\end{tabular}
\end{table}

\begin{figure}[t]
\centering
\includegraphics[width=0.85\textwidth]{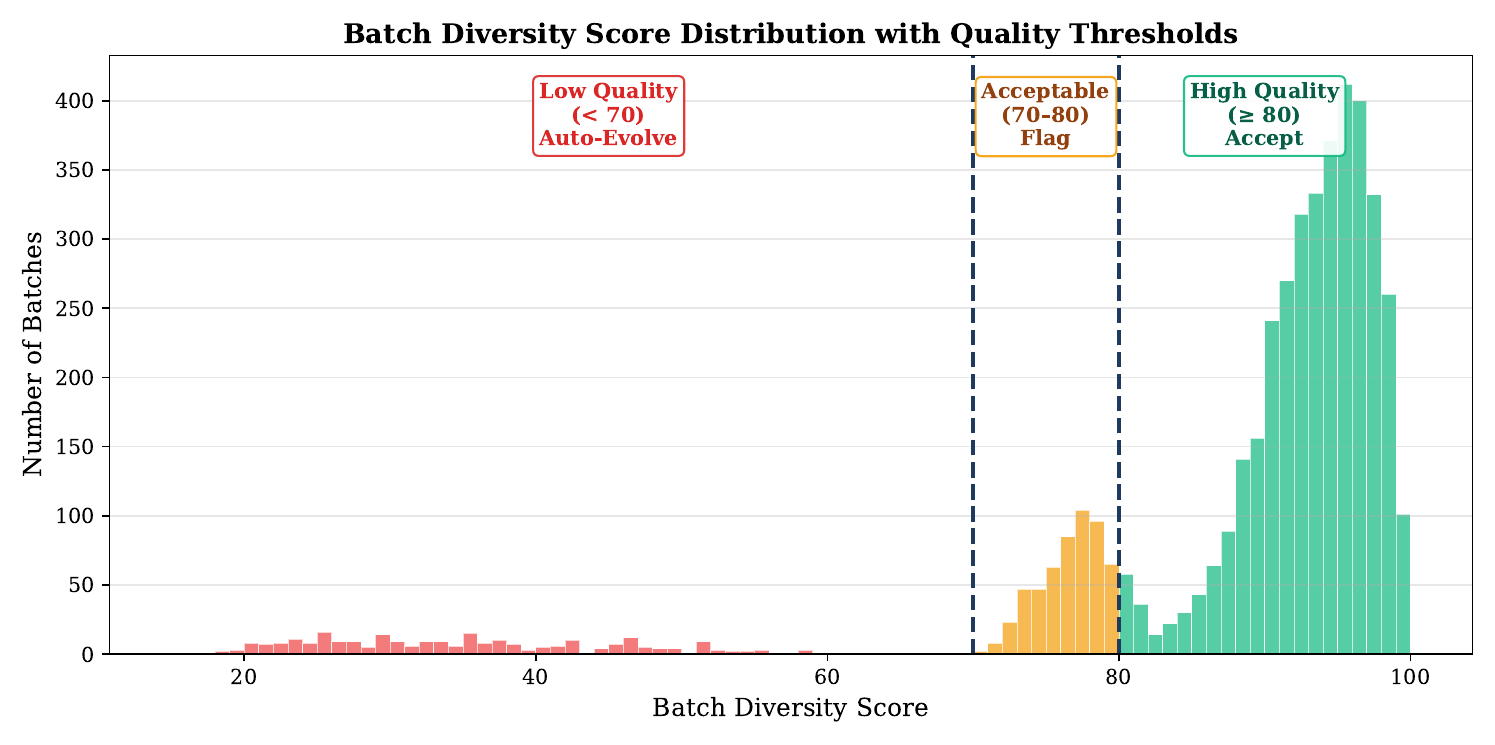}
\caption{Distribution of diversity scores across 4,500 human-validated batches. Threshold at 70 triggers auto-evolve regeneration; threshold at 80 marks high-quality batches suitable for training without modification.}
\label{fig:diversity_distribution}
\end{figure}

Figure~\ref{fig:diversity_distribution} shows the distribution of diversity scores, with thresholds at 70 (triggers auto-evolve) and 80 (high-quality batches accepted without modification).

\section{Data Generation Pipeline}
\label{sec:generation}

Our pipeline transforms 632K taxonomy tuples into 14.3M quality-filtered samples through four stages: pre-generation research (establishing optimal parameters), conditional generation with auto-evolution (producing 26.4M raw samples, merged to 20.9M), two-layer semantic deduplication (reducing to 16.9M), and LLM-as-Judge quality filtering (reducing to 14.3M). The first two stages produce and refine samples; the latter two are filtering stages that remove duplicates and low-quality samples respectively.

\subsection{Sample Allocation Strategy}

Given $T$ taxonomy tuples, the total raw sample count is:
\begin{equation}
N_{\text{raw}} = T \times \sum_{m \in \mathcal{M}} \left(n_{\text{unsafe}}^{(m)} + n_{\text{safe}}^{(m)}\right) = T \times (20 + 8 + 3) = 31T
\end{equation}

With $T \approx 632{,}000$ unique tuples, this yields approximately 19.6M base samples. Including Evol-Instruct augmentation and auto-evolved variants, the raw SDG output totals 26,422,384 samples (Bank: 9,224,426; Insurance: 8,535,183; Financial Services: 8,662,775).

The allocation strategy is governed by three principles:

\paragraph{Complexity-Adjusted Allocation.} Generation cost increases super-linearly with mode complexity. Input mode generates only user prompts; output mode generates prompt + response; multi-turn generates $k$ coherent turns with quadratic context growth.

\paragraph{Diminishing Marginal Diversity.} Beyond a threshold, additional samples yield diminishing diversity returns. Our batch size experiments confirm this: input mode plateaus at $n \approx 20$, output at $n \approx 10$, multi-turn at $n \approx 5$.

\paragraph{Unsafe-Heavy Ratio.} We generate more unsafe than safe samples per tuple (3:1 for input, 5:3 for output, 2:1 for multi-turn) because safety classifiers need rich coverage of violation patterns.

\subsection{Auto-Evolve Process}

A key innovation is the \textit{chained auto-evolve} mechanism, which automatically detects and remediates low-diversity batches during generation.

\paragraph{Trigger Condition.} After generating the initial batch for each (tuple, mode) pair, we compute the 5-component diversity score. If the score falls below 70, the batch is flagged for evolution-based retry:
\begin{equation}
\text{AutoEvolve}(\text{tuple}, m) = \mathbb{1}\left[D(\text{batch}) < 70\right]
\end{equation}

\paragraph{Evolution Operators.} Retry batches are regenerated using Evol-Instruct \cite{xu2023wizardlm} evolution. The evolution applies operators including: add constraints, deepen reasoning, concretize abstractions, increase complexity, and complicate inputs.

\paragraph{Sample Replacement.} Evolved samples are marked with an \texttt{\_auto\_evolved\_replacement} flag. During the merge step, records are grouped by (\texttt{tuple\_id}, \texttt{mode}, \texttt{target\_label}). For each group, if evolved replacements exist, they take priority over originals.

\paragraph{Merge Impact.} The merge step reduces 26,422,384 raw samples to 20,909,187 (20.9\% reduction, removing $\sim$5.5M records). This reduction comes from: (1) discarding original samples superseded by auto-evolved replacements, and (2) resolving overlap from parallel generation shards.

\section{Two-Layer Deduplication}
\label{sec:deduplication}

After merging, we perform global deduplication to remove semantically similar content generated from related taxonomy tuples. Prior work has demonstrated that deduplication significantly improves language model quality \cite{lee2022deduplicating,abbas2023semdedup}.

\subsection{Deduplication Architecture}

Our pipeline implements two complementary layers:

\paragraph{Layer 1: LSH + SimHash (Lexical).} Locality-Sensitive Hashing enables sub-quadratic lexical duplicate detection. Each sample's text is converted to a 64-bit SimHash fingerprint, split into 8 bands, and bucketed. Candidate pairs with Hamming distance $\leq 5$ are confirmed as duplicates. Complexity: $O(N)$ for bucketing, $O(|\text{candidates}|)$ for verification.

\paragraph{Layer 2: FAISS + MiniLM (Semantic).} To catch paraphrases missed by lexical methods, we apply embedding-based semantic deduplication. Each sample is encoded into a 384-dimensional vector using \texttt{all-MiniLM-L6-v2}. We build a FAISS \texttt{IndexIVFFlat} with $\sqrt{N}$ Voronoi cells for approximate nearest neighbor search. Pairs with cosine similarity $\geq 0.85$ are flagged as semantic duplicates. Complexity: $O(N)$ for embedding, $O(N \log N)$ for FAISS index construction and search.

\paragraph{Layer Interaction.} The semantic layer (FAISS) subsumes the lexical layer: any text pair with near-identical surface form will necessarily have high embedding similarity. The semantic layer additionally catches paraphrases that lexical methods miss entirely, providing the bulk of the deduplication signal.

\subsection{Deduplication Results}

Table~\ref{tab:dedup_impact} presents the deduplication impact by domain and mode.

\begin{table}[t]
\centering
\caption{Deduplication Impact by Domain and Mode (FAISS, cosine $\geq 0.85$)}
\label{tab:dedup_impact}
\begin{tabular}{@{}llrrr@{}}
\toprule
\textbf{Domain} & \textbf{Mode} & \textbf{Pre-Dedup} & \textbf{Post-Dedup} & \textbf{\% Removed} \\
\midrule
Banking & Input & 4,337,957 & 3,351,116 & 22.7\% \\
Banking & Output & 1,739,168 & 1,032,734 & 40.6\% \\
Banking & Multi-Turn & 1,083,563 & 822,789 & 24.1\% \\
\midrule
Insurance & Input & 4,166,222 & 3,717,065 & 10.8\% \\
Insurance & Output & 1,666,469 & 1,284,466 & 22.9\% \\
Insurance & Multi-Turn & 1,037,367 & 883,066 & 14.9\% \\
\midrule
Financial Svcs & Input & 4,172,360 & 3,760,956 & 9.9\% \\
Financial Svcs & Output & 1,668,864 & 1,252,687 & 24.9\% \\
Financial Svcs & Multi-Turn & 1,037,217 & 889,736 & 14.2\% \\
\midrule
\textbf{Total} & & \textbf{20,909,187} & \textbf{16,994,615} & \textbf{18.7\%} \\
\bottomrule
\end{tabular}
\end{table}

\subsection{Domain and Mode Analysis}

\paragraph{Output Mode Exhibits Highest Duplication.} Output mode shows 22.9--40.6\% deduplication across all domains, significantly higher than input (9.9--22.7\%) or multi-turn (14.2--24.1\%). This aligns with theoretical analysis: assistant responses to similar queries tend to converge on standard refusal patterns or compliance templates. Banking output exhibits the highest rate (40.6\%), likely due to regulatory response templates.

\paragraph{Domain-Specific Patterns.} Banking shows higher duplication than Insurance and Financial Services across all modes. Banking tuples share more overlapping terminology (account, transaction, transfer), leading to convergent outputs. Financial Services input shows the lowest duplication (9.9\%), reflecting greater lexical diversity in financial services prompts.

\paragraph{Multi-Turn Moderate Duplication.} Multi-turn conversations show 14.2--24.1\% deduplication. The additional context from multiple turns creates more unique surface forms even when underlying intent overlaps, reducing duplicate detection rates.

\paragraph{Paraphrase Detection Examples.} The semantic layer catches paraphrases that lexical methods miss:
\begin{itemize}
\item ``How can I transfer funds internationally?'' $\leftrightarrow$ ``What's the process for overseas wire transfers?''
\item ``Help me minimize my tax obligations'' $\leftrightarrow$ ``What strategies reduce my tax burden legally?''
\end{itemize}

\section{LLM-as-Judge Quality Evaluation}
\label{sec:quality}

After deduplication, all 16.9M samples undergo LLM-as-Judge quality evaluation to assess behavioral correctness and tuple fidelity. Samples failing quality thresholds are filtered, reducing the dataset from 16.9M to the final 14.3M training corpus.

\subsection{Two-Stage Evaluation Framework}

\paragraph{Stage 1: Behavioral Assessment.} The judge LLM evaluates whether the generated sample exhibits the intended safety behavior specified by the target label. For unsafe samples, this means checking if the content genuinely violates safety guidelines. For safe samples, this verifies appropriate handling of the scenario. The judge outputs a binary assessment (PASS/FAIL) with verbalized confidence (0--100).

\paragraph{Stage 2: Fidelity Checking.} The judge evaluates whether the sample accurately reflects the constraints specified in the taxonomy tuple: correct domain, appropriate risk level, matching framing style, suitable for the specified channel, and consistent with user sophistication. This produces a fidelity score (0--5) with group-wise pass/fail indicators.

\paragraph{Confidence Delta.} For each sample, we compute:
\begin{equation}
\Delta_{\text{conf}} = \text{conf}_{\text{stage2}} - \text{conf}_{\text{stage1}}
\end{equation}

Positive $\Delta_{\text{conf}}$ indicates the fidelity check increased confidence in the behavioral assessment. Negative $\Delta_{\text{conf}}$ suggests potential misalignment between content and tuple specification, flagging samples for downstream review.

\subsection{Quality Labels for Downstream Filtering}

The judge produces a confidence score and fidelity score for each sample. We apply empirically determined thresholds to filter the dataset into three regions:

\begin{itemize}
\item \textbf{High-confidence training:} Filter to samples with $\text{conf} > 70$ and $\Delta_{\text{conf}} \geq 0$
\item \textbf{Balanced training:} Include samples with $50 \leq \text{conf} \leq 70$ and fidelity $\geq 3$
\item \textbf{Edge case focus:} Target samples with $\text{conf} < 50$ for exclusion from training
\end{itemize}

This three-region approach reduces the post-deduplication dataset from 16.9M to \textbf{14.3M} samples (16.5\% reduction). The filtering disproportionately removes low-quality safe-labeled samples, shifting the label distribution slightly toward unsafe (75.3\% vs.\ 72.8\% pre-filtering).

\section{Limitations and Future Work}
\label{sec:limitations}

\paragraph{Domain Scalability.} Our methodology is tailored to financial AI safety. Extending to healthcare, legal, or government domains requires domain-specific taxonomies and validation of diversity metrics against domain expert ratings.

\paragraph{Multilingual Extension.} The current pipeline generates English-only data. Global deployment requires multilingual generation with jurisdiction-specific regulatory constraints and cultural context.

\paragraph{Computational Cost Optimization.} Generating 26.4M samples, computing 16.9M embeddings for deduplication, and running LLM-as-Judge evaluation on 16.9M samples is computationally expensive. Future work should explore adaptive batch sizing, hierarchical deduplication, and distilled judge models.

\paragraph{Alternative Deduplication Methods.} Our FAISS-based approach uses fixed cosine similarity thresholds. Cluster-based methods, diversity-aware sampling, or learned duplicate classifiers may improve precision-recall tradeoffs.

\paragraph{Quality Scoring Improvements.} Single-judge evaluation may miss subtle quality issues. Multi-judge ensembles, calibrated confidence scoring, and learned fidelity models could enhance quality assessment.

\paragraph{Temporal Dynamics.} Financial regulations evolve over time. Continuous learning systems that track regulatory changes and update the taxonomy accordingly would maintain dataset relevance.

\paragraph{Downstream Evaluation.} While we validate diversity metrics against human ratings, comprehensive evaluation requires training models on this data and measuring downstream task performance, conducting ablation studies on pipeline components, and deploying in real-world financial AI systems.

\paragraph{Explainability.} Current LLM-as-Judge assessments provide limited reasoning. Structured explanations, counterfactual examples, and audit trails would improve transparency for regulatory compliance.

\paragraph{Bias and Fairness.} Synthetic data may inherit biases from the generation model. Systematic auditing for demographic bias, fairness constraints in generation, and alignment with regulatory fairness requirements are essential for production deployment.

\section{Conclusion}
\label{sec:conclusion}

We presented a comprehensive methodology for large-scale synthetic dataset generation, producing 14.3 million quality-filtered training examples from a structured taxonomy of 632,000 scenario specifications. Our four-stage pipeline addresses critical challenges in synthetic data quality, diversity, and realism through systematic pre-generation research, conditional auto-evolution, two-layer semantic deduplication, and LLM-as-Judge quality filtering.

Key contributions include: (1) empirical batch size optimization revealing mode-specific diversity saturation thresholds, (2) validated 5-component diversity scoring with strong correlation to human judgments, (3) conditional auto-evolve mechanism reducing raw output by 20.9\%, (4) domain-specific deduplication analysis showing 9.9--40.6\% redundancy patterns, and (5) complete pipeline reducing 26.4M raw samples to 14.3M high-confidence examples through principled quality control (45.8\% total reduction).

Our methodology provides a reproducible framework for domain-specific synthetic data generation with rigorous quality control, applicable to any specialized domain requiring large-scale training data with comprehensive coverage of edge cases and failure modes.

\bibliographystyle{splncs04}
\bibliography{references}

\end{document}